\documentclass{article}

\usepackage{spconf,amsmath,epsfig}

\usepackage{amssymb,algorithm}
\usepackage[noend]{algpseudocode}
\usepackage[super]{nth}
\usepackage{graphicx,multirow}
\usepackage{enumitem}
\usepackage[dvipsnames]{xcolor}

\begin{document}\sloppy

\nocite{*}

\def\x{{\mathbf x}}
\def\L{{\cal L}}

\title{Learning Goal-Oriented Visual Dialog Agents: \\Imitating And Surpassing Analytic Experts}
%
\name{Yen-Wei Chang, Wen-Hsiao Peng}
\address{National Chiao Tung University, Taiwan \\
\{changyw, wpeng\}@cs.nctu.edu.tw}

\maketitle

\begin{abstract}
This paper tackles the problem of learning a questioner in the goal-oriented visual dialog task. Several previous works adopt model-free reinforcement learning. Most pretrain the model from a finite set of human-generated data. We argue that using limited demonstrations to kick-start the questioner is insufficient due to the large policy search space. Inspired by a recently proposed information theoretic approach, we develop two analytic experts to serve as a source of high-quality demonstrations for imitation learning. We then take advantage of reinforcement learning to refine the model towards the goal-oriented objective. Experimental results on the GuessWhat?! dataset show that our method has the combined merits of imitation and reinforcement learning, achieving the state-of-the-art performance.
\end{abstract}

%
\section{Introduction}
\label{sec:intro}
Research on goal-oriented visual dialogue~\cite{visdial,guesswhat_game} has recently attracted lots of attention. Unlike the conventional VQA~\cite{VQA}, where the robot answerer has to answer any question related to an input image raised by a human even if the question itself is ambiguous or indefinite, the goal-oriented visual dialogue extends the question-answering interactions to multiple rounds, turning the robot into also a questioner which can retrieve more information from the human. This challenging task calls for the robot's ability to reason over both visual scenes and textual dialogue history.

\begin{figure}[ht]
\centering
\includegraphics[width=8.6cm]{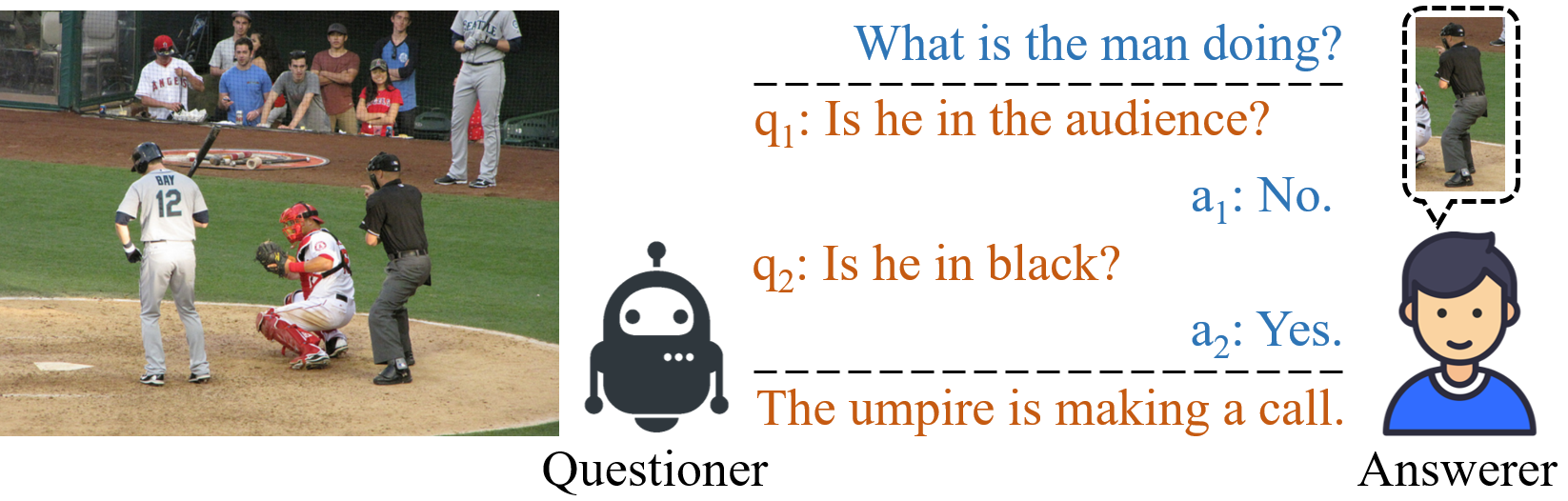}
\vspace{-0.8cm}
\caption{Illustration of a real-life GuessWhat?! scenario where a boy who is new to the baseball game asks an ambiguous question ``What is the man doing?'' and the robot tries to figure out the man the boy is talking about by asking back discriminative questions.}
\label{fig:intro}
\vspace{-0.35cm}
\end{figure}

For developing such goal-oriented visual dialogue systems, GuessWhat?! is one commonly used dataset. It specifies a goal-oriented 2-player object guessing game, where a questioner (a robot) tries to figure out the target object in an image chosen by an answerer (normally a human). The questioner must learn to ask critical questions to the answerer in order to get useful information for identifying the target object. A real-life GuessWhat?! scenario is illustrated in Fig.~\ref{fig:intro}.

Some previous works~\cite{end_to_end_gw,Zhao2018ImprovingGV} address this GuessWhat?! task by applying reinforcement learning (RL) to train the questioner. The process often involves pretraining the questioner from a finite set of human-generated data. However, due to the large policy search space, they struggle with reaching satisfactory performance. 

Lately, an information theoretic approach, Answerer in Questioner's Mind (AQM)~\cite{NIPS2018_7524}, is proposed. Unlike the RL-based methods, AQM constructs an approximate model of the answerer so that it can be queried by the questioner to predict what question would induce an answer that minimizes the uncertainty about the target object. Although AQM outperforms similar prior works by a large margin, its performance is highly variable depending on how accurate the approximate model is. 

In this work, we leverage the merits of both approaches. We make use of the probabilistic framework in~\cite{NIPS2018_7524} to devise an Information Gain Expert and a Target Posterior Expert. These experts are queried to provide virtually unlimited expert demonstrations for pretraining the questioner. Since they are not perfect experts, we refine the model using the REINFORCE algorithm to discover a even better policy. Extensive experiments confirm the superiority of our method over several state-of-the-art baselines in terms of prediction accuracy and robustness to the model approximation error.

\section{Related Work}\label{sec:related_work}
\textbf{Goal-Oriented Visual Dialogue.} GuessWhat?!~\cite{guesswhat_game} is a collaborative 2-player visual grounded object discovery game. The game begins with presenting an image $\mathcal{I}$ of a rich visual scene containing $M$ objects $C=\{c_m\}_{m=1}^M$ to both players, the questioner and the answerer. The answerer first picks in mind an object $c^* \in C$, which is unknown to the questioner. The questioner then tries to identify the target object $c^*$ by asking a yes-no question $q_j$ to the answerer, who responds with an answer $a_j$ of yes, no or not applicable. The process continues for $T$ iterations, forming a QA history $h_{1:T}=\{q_j,a_j\}_{j=1}^T$. Once the questioner finishes asking questions, the list of candidate objects $C$ is finally shown to the questioner. The questioner then makes a guess of the target object $c^*$ based on $h_{1:T}$. The game is considered successful if the target object $c^*$ is correctly identified .

\noindent
\textbf{Model-Free Reinforcement Learning (RL).} To train a questioner for solving the GuessWhat?! game,~\cite{end_to_end_gw,Zhao2018ImprovingGV} construct an ``oracle'' network to mimic the answerer's behavior, regard it as part of the environment in the reinforcement learning setup and then apply the REINFORCE algorithm (or Monte Carlo Policy Gradient). The questioner learns to ask critical questions that help identify the target object by interacting with the oracle. In the training process, the questioner is rewarded when it successfully identifies the target object after few QA iterations. The term ``model-free'' refers to the fact that the questioner has no knowledge of the oracle model. Typically, for converging quickly to a good policy, the questioner is pretrained supervisedly based on human-generated data. 

\noindent
\textbf{Answerer in Questioner's Mind (AQM).} AQM~\cite{NIPS2018_7524} adopts an information-theoretic approach to design the questioner. Unlike the RL-based methods, AQM has an objective of asking a question in each QA round that would induce an answer minimizing the uncertainty about the target object. This notion is formalized as asking a question that maximizes the conditional mutual information between the target object and the answer. The process involves the modeling of a posterior distribution $p(c|q_t, h_{1:t-1}, \mathcal{I})$ over the candidate objects $C$ at each iteration $t$ given the question $q_t$, the previous QA history $h_{1:t-1}=\{q_j,a_j\}_{j=1}^{t-1}$ and the image $\mathcal{I}$ as well as the modeling of the oracle's behavior, characterized by the conditional distribution $p(a_t|c^*,q_t,h_{1:t-1},\mathcal{I})$\footnote{In AQM, the oracle is implemented in such a way that the answer $a_t$ is conditionally independent of the QA history $h_{1:t-1}$ and $\mathcal{I}$ given the target object $c^*$ and the current question $q_t$; that is, $p(a_t|c^*,q_t,h_{1:t-1}, \mathcal{I})=p(a_t|c^*,q_t)$.} of its response $a_t$, where $c^*$ is the target object. Since the true oracle in reality is not accessible, AQM calls for an approximate model $\tilde{p}(a_t|c^*,q_t,h_{1:t-1},\mathcal{I})$ of the oracle, explaining the origin of its name ``Answerer in Questioner's Mind''. Currently, AQM uses a selection of predefined questions instead of an open question set. It achieves the state-of-the-art performance in the GuessWhat?! game. 

\noindent
\textbf{Imitation Learning (IL). } In IL~\cite{DBLP:conf/icra/KahnZLA17,pmlr-v15-ross11a,Peng:2018:DED:3197517.3201311,NIPS2014_5421}, the learner tries to achieve the best performance by mimicking the expert's moves. In learning a questioner for the GuessWhat?! game, it is important to pretrain the model (the learner) supervisedly based on human-generated data, which is a form of imitation learning. However, applying direct behavior cloning by learning supervisedly from a finite set of the expert's demonstrations is impractical since the learner only learns how to behave in states that have been visited by the expert. It may fail to generalize to states never seen by the expert, leading to the so called cascading error in the long run, i.e. the state trajectories traversed by the learner may end up deviating significantly from those traversed by the expert. To mitigate the cascading error,~\cite{pmlr-v15-ross11a} suggests that the learner should also learn the expert's actions over states visited by the learner itself in addition to those visited by the expert. However, this requires an expert, e.g. a human subject, to be queried whenever needed, which is often impractical. 

\begin{figure}[t]
\centering
\includegraphics[width=8.6cm]{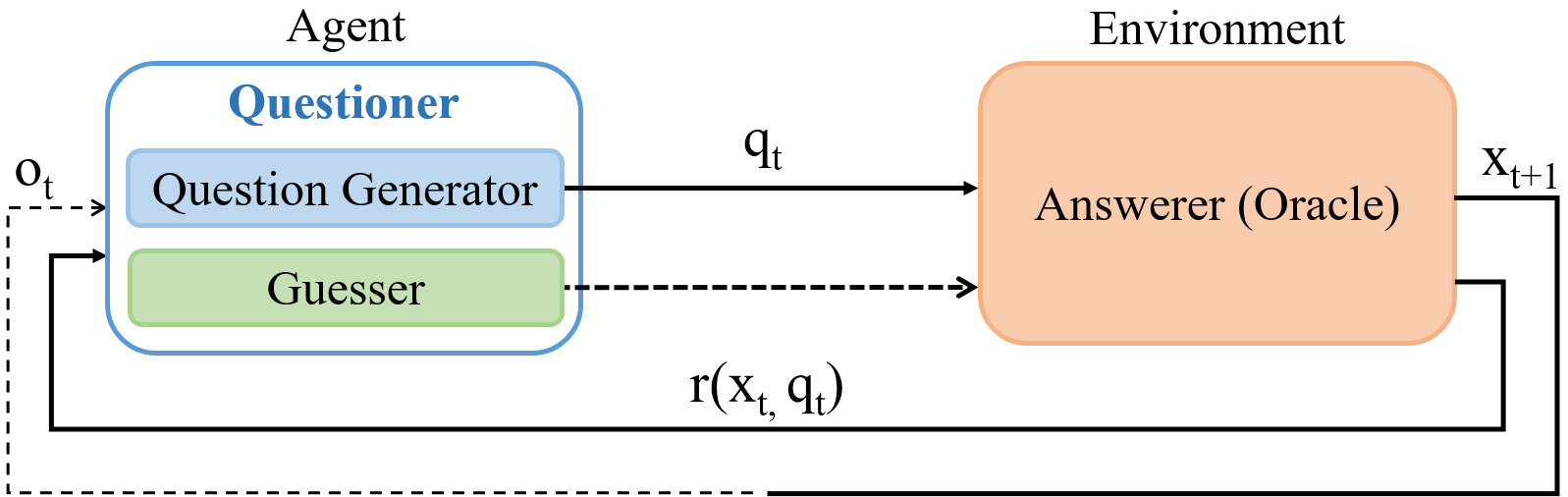}
\vspace{-0.7cm}
\caption{The MDP formulation of the GuessWhat?! game.}
\label{fig:mdp}
\vspace{-0.35cm}
\end{figure}

%
\section{Method}\label{sec:method}
To overcome the problems faced by these previous works, our proposed method obtains a better questioner by learning from analytic experts, which provide virtually unlimited demonstrations, and by taking advantage of RL to discover a even better policy than the experts', which suffer inherently from imperfect modeling of the oracle. In this session, we give a formal treatment of the proposed method, starting from formalizing the GuessWhat?! game as a Markov decision process to learning the questioner by imitation learning and policy gradient, and ending with implementation details. 

\subsection{GuessWhat?! as A Markov Decision Process}
\label{sec:gw_game}
We formalize the GuessWhat?! game as a Markov decision process (MDP). 
We start by viewing the questioner as a two-part task comprising (1) a question generator, which is to ask a question $q_{t}$ in the $t$-th QA round based on the previous dialog history $h_{1:t-1}=\{q_j,a_j\}_{j=1}^{t-1}$ and the given image $\mathcal{I}$, and (2) a guesser, which makes a guess of the target object $c^*$ by observing the list of candidate objects $C$, the dialog history $h_{1:t-1}$ and the image $\mathcal{I}$. From this perspective, the interaction between the questioner and the oracle (or the answerer) in $T$ rounds of QA can be formalized as a MDP $(X, Q, P, R)$: 

\begin{algorithm}[!t]
\caption{Training of Question Generator}\label{algo:1}
\begin{algorithmic}[1]
  \Require 
    \Statex $Q \gets Q$-sampler
    \Statex A pretrained oracle $p(a_t|c, q_t, h_{1:t-1}, \mathcal{I})=p(a_t| q_t, x_t)$
    \Statex $\pi_\theta \gets$ Random Initialization
    \Statex $\pi_{IGE}^*$, $\pi_{TPE}^* \gets$ Experts
    
  \Procedure{Imitation Learning (DAgger~\cite{pmlr-v15-ross11a})}{}
     \State Initialize dataset $\mathcal{D} \gets \emptyset$
     
     \For {$i=1$ \textbf{to} $N$}
       \State $x_1 \gets $ Random $list(c^*, \mathcal{I})$
       \For {$t=1$ \textbf{to} $T$}
         \State $o_t = x_t.remove(c^*)$
         \State $\hat{\pi}_i=\beta_i\pi_\theta(q_t|o_t) + (1-\beta_i)\pi^*(q_t|x_t)$
         \State Sample $\hat{q_t} \sim \hat{\pi}_i$
         \State Sample $q_t^* \sim \pi^*(q_t|x_t)$ {\it \# expert's move}
         \State Aggregate $pair(o_t, q_t^*)$ into dataset $\mathcal{D}$
         \State Sample $a_t \sim p(a_t|\hat{q_t}, x_t)$
         \State State evolves $x_{t+1}=x_t.append(\hat{q_t}, a_t)$
      \EndFor
      \State Train classifier $\pi_\theta$ on $\mathcal{D}$
    \EndFor
  \EndProcedure
  
  \Require
  \Statex An approximate oracle $\tilde{p}(a_t|q_t, c, \mathcal{I})$
  \Statex A pretrained $\pi_\theta$ from imitation learning
  \Procedure{REINFORCE}{}
    \For {$i=1$ \textbf{to} $N$}
       \State $x_1 \gets $ Random $list(c^*, \mathcal{I})$
       \For {$t=1$ \textbf{to} $T$}
         \State $o_t = x_t.remove(c^*)$
         \State Sample $q_t \sim \pi_\theta(q_t|o_t)$
         \State Sample $a_t \sim p(a_t|q_t, x_t)$
         \State State evolves $x_{t+1}=x_t.append(q_t, a_t)$
      \EndFor
      \State Episodic reward $R(\tau)=0$
      \If {$\mathop{\arg\max}_{c} p(c|h_{1:T},\mathcal{I}) = c^*$}
      \State $R(\tau)=1$
      \EndIf
      \State Compute $\nabla_\theta J(\theta)$ (cf. Eq.~\eqref{eq:pg})
      \State Update $\theta=\theta+\alpha\nabla_\theta J(\theta)$
    \EndFor
  
  \EndProcedure
\end{algorithmic}
\end{algorithm}

\begin{itemize}
\itemsep=-4pt
\item $X$ represents the state of the MDP. We define the state  $x_t=(h_{1:t-1},c^*, \mathcal{I})$ in the $t$-th QA round to be composed of the previous QA history $h_{1:t-1}$, the target object $c^*$, and the image $\mathcal{I}$.

\item $Q$ denotes the action taken by the questioner (or the agent in RL language) in state $X$. 
In the present case, the action is the question $q_t$ output by the question generator. 

\item $P$ is the state transition probability $P_{XX'}^Q=p(X'=x_{t+1}|X=x_t,Q=q_t)$, which is uniquely determined by the oracle's behavior, $p(a_t|c^*, q_t, h_{1:t-1}, \mathcal{I})$.  

\item $R$ is the reward signal given to the questioner. We denote the immediate reward for the questioner taking action $q_t$ in state $x_t$ in the $t$-th QA round by $r(x_t, q_t)$. Since the game terminates after $T$ QA rounds, the interaction between the questioner and the oracle forms a state-action-reward sequence $\tau=(x_1, q_1, r(x_1, q_1), ...,x_{T}, q_{T},r(x_T, q_T))$. The cumulative reward is seen to be $R(\tau)=\sum_{t=1}^T r(x_t, q_t)$.
\end{itemize}

It is worth noting that the state $X$ is not fully visible. According to the rules of the game, the questioner does not have access to the target object $c^*$. In this sense, the MDP is partially observable. We then have the question generator governed by $p(q_t|x_t)=p(q_t|o_t)$; in other words, the question $q_t$ has a distribution that depends solely on the visible observations $o_t=(h_{1:t-1}, \mathcal{I})$. As will be seen shortly, we further implement the question generator $p(q_t|o_t) \approx \pi_\theta(q_t| o_t)$ by a neural network $\pi_\theta$ parameterized by $\theta$. 

\subsection{Learning}
The learning of the question generator is done in two sequential phases, the imitation learning (IL) phase and the reinforcement learning (RL) phase. The former is to pretrian the question generator by imitating experts, while the latter is to discover a better policy than the experts'. In the IL phase, we get access to the true oracle to construct two analytic experts to solve the game and pretrain the question generator through imitating these experts. In the RL phase, we further equip it with an approximate oracle and apply RL to direct the learning objective towards successful identification of the target object. The following expand on these two learning phases.

\subsubsection{Imitation Learning from Analytic Experts}
\label{sec:ILExperts}
For IL, we assume that we have full knowledge of the oracle (or the environment in RL language). That is, we have control over the target object $c^*$ and access to the oracle's behavior $p(a_t|c^*, q_t, h_{1:t-1}, \mathcal{I})$, which implies that the state transition probability $P$ of the MDP is known. This allows us to produce virtually unlimited demonstrations by having any expert or planner interact with the oracle. To automate the generation process, we introduce two analytic experts as follows.

\noindent
\textbf{Information Gain Expert (IGE).} Inspired by AQM ~\cite{NIPS2018_7524}, IGE calculates in each QA round the information gain $I[C,A_t|q_t, h_{1:t-1},\mathcal{I}]$, defined as the conditional mutual information between the object $C$ and the answer $A_t$ given the question $q_t$, the QA history $h_{1:t-1}$ and the image $\mathcal{I}$, for every question $q_t$ in $Q$ and chooses the one $q_t^*$ with the maximum information gain to be its action:
\vspace*{-0.2\baselineskip}
\begin{equation}
\label{eq:IG}
\begin{aligned}
q_t^*= & \arg\max_{q_t \in Q}    I[C,A_t|q_t,h_{1:t-1},\mathcal{I}] \\ 
     = & \arg\max_{q_t \in Q}   \sum_{a_t,c}  p(c|q_t,h_{1:t-1},\mathcal{I}) \times p(a_t|c,q_t,h_{1:t-1},\mathcal{I}) \\
       & \qquad \qquad \times                                  \ln{\frac{p(a_t|c,q_t,h_{1:t-1},\mathcal{I})}{p(a_t|q_t,h_{1:t-1},\mathcal{I})}}, 
\end{aligned}
\end{equation}
where the posterior distribution $p(c|q_t,h_{1:t-1},\mathcal{I})$ is iteratively updated following
\vspace*{-0.2\baselineskip}
\begin{equation}
\label{eq:poscal}
\begin{aligned}
& p(c|q_t,h_{1:t-1},\mathcal{I}) \\
& \propto p(c|\mathcal{I}) \times \left(\prod_{j=1}^{t-1} p(a_j|c, q_j, h_{1:j-1},\mathcal{I})p(q_j|c, h_{1:j-1},\mathcal{I})\right) \\
& \qquad \qquad \times p(q_t|c,h_{1:t-1},\mathcal{I}) \\
& \propto p(c|\mathcal{I}) \times \prod_{j=1}^{t-1} p(a_j|c, q_j, h_{1:j-1},\mathcal{I}).
\end{aligned}
\end{equation}
The last proportionality is because the question generator has no access to $c$, i.e. $p(q_t|c,h_{1:t-1},\mathcal{I})=p(q_t|h_{1:t-1},\mathcal{I})$, a constraint that has been discussed in Section \ref{sec:gw_game}.

It is interesting to see that evaluating the information gain in Eq.~\eqref{eq:IG} and the posterior distribution in Eq.~\eqref{eq:poscal} involves the oracle model, $p(a_j|c, q_j, h_{1:j-1},\mathcal{I})$. Since the true oracle is unavailable in practice, an approximate model $\tilde{p}(a_j|c, q_j, h_{1:j-1},\mathcal{I})$ of the oracle is used instead. It is the approximation error that crucially affects the performance of IGE (and also AQM) at test time. 


\begin{figure}[t]
\begin{minipage}[b]{.49\linewidth}
  \centering
  \centerline{\epsfig{figure=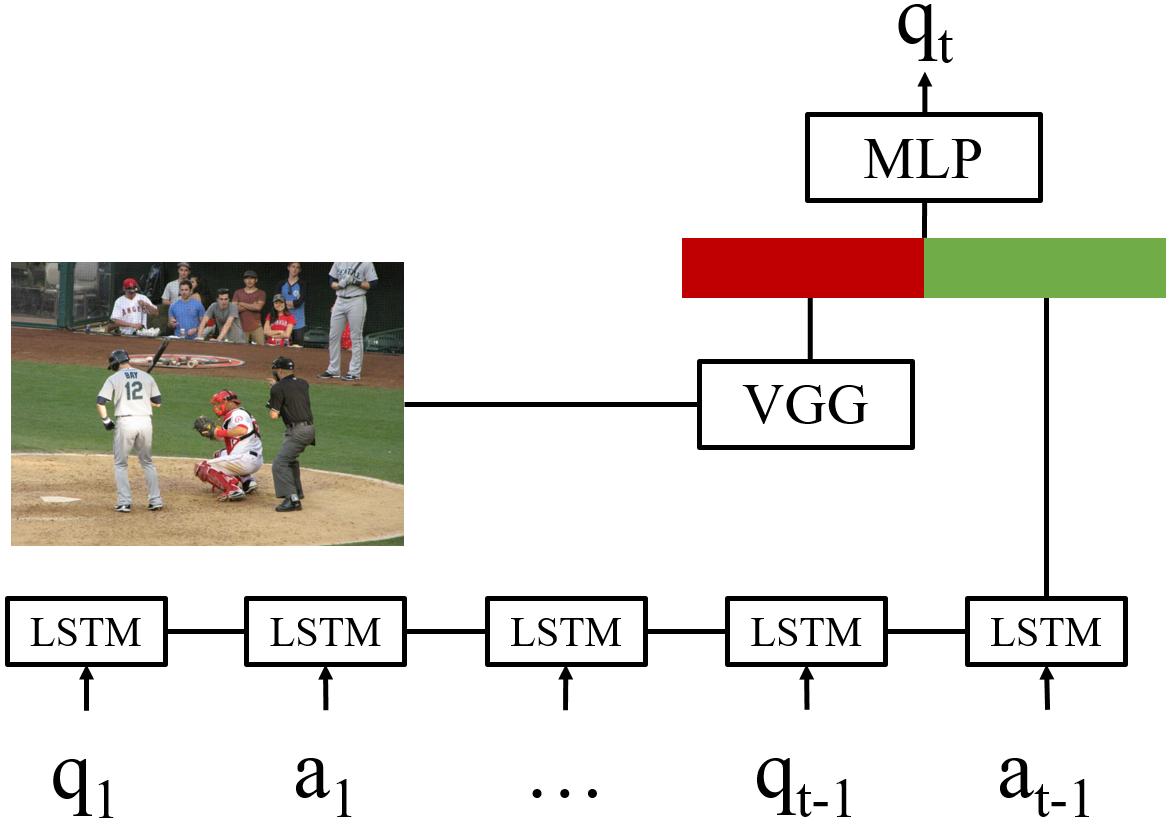,width=4.2cm}}
  \vspace{0cm}
  \centerline{(a) Question generator model.}\medskip
\end{minipage}
\hfill
\begin{minipage}[b]{0.49\linewidth}
  \centering
  \centerline{\epsfig{figure=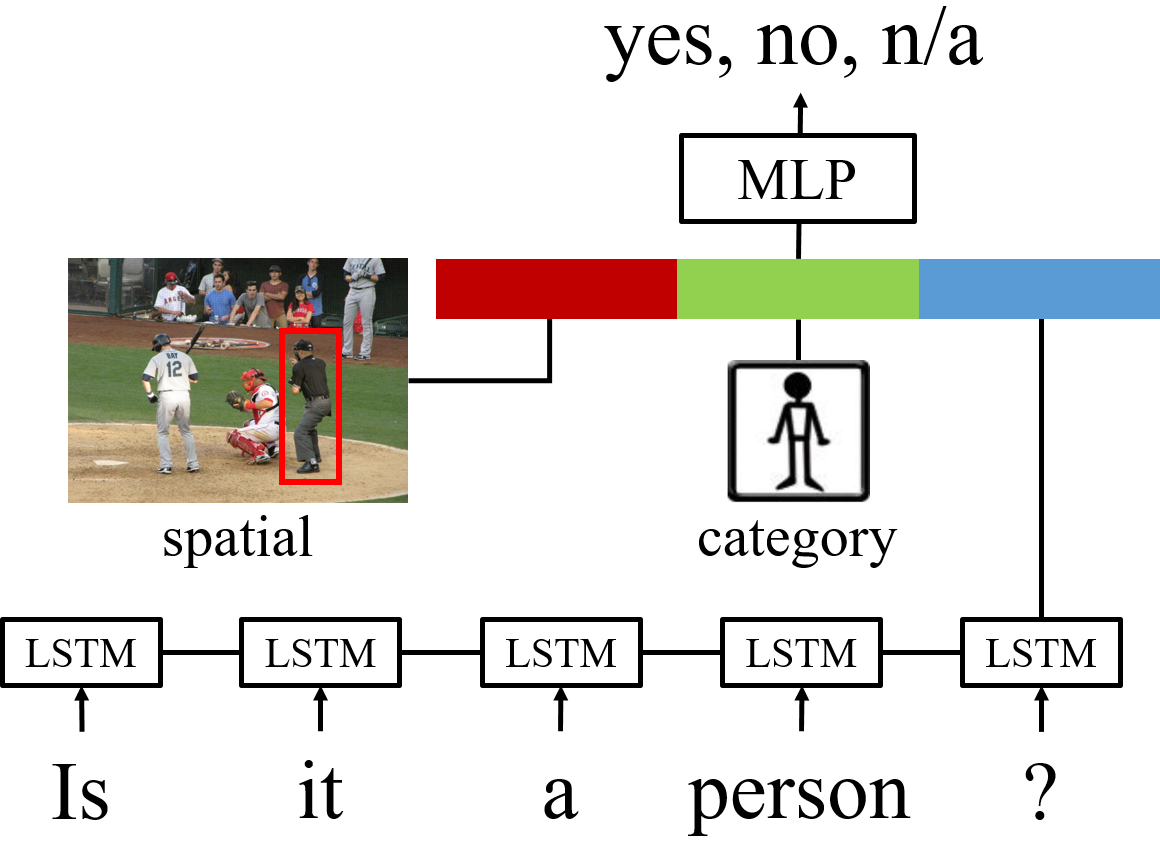,width=4.2cm}}
  \vspace{0cm}
  \centerline{(b) Oracle model.}\medskip
\end{minipage}
  \vspace{-0.55cm}
\caption{The question generator and the oracle in our method.}
\label{fig:nn}
  \vspace{-0.3cm}
\end{figure}

\noindent
\textbf{Target Posterior Expert (TPE).} TPE makes use of the knowledge about the target object $c^*$ to choose a question $q_t^*$ that maximizes the following posterior probability:
\begin{equation}
q_t^*=\arg\max_{q_t \in Q} p(c^*|h_{1:t},\mathcal{I}),
\end{equation}
where $p(c^*|h_{1:t},\mathcal{I}) \!\propto\! p(c^*|\mathcal{I})\times \prod_{j=1}^{t} p(a_j|c^*, q_j, h_{1:j-1},\mathcal{I})$.

Note that the posterior distribution $p(c|h_{1:t},\mathcal{I})$ is also queried by the guesser inside the questioner to make a guess of the target object by $\arg \max_c p(c|h_{1:t},\mathcal{I})$.

To sum up, both IGE and TPE are not perfect due to the need to incorporate an approximate oracle model for estimating the information gain and for making a guess. Moreover, they are greedy in that they amount to performing one-step dynamic programming. Recognizing their limitations, we put them into use in the IL framework of DAgger~\cite{pmlr-v15-ross11a} for the sheer purpose of pretraining our question generator (see the imitation learning part of Algorithm \ref{algo:1}). 

\vspace{-0.05cm}
\subsubsection{Reinforcement Learning with Policy Gradient}
The RL phase is to refine the pretrained question generator to optimize for successful identification of the target object. We view the questioner, including the question generator and the guesser, as the agent and the true oracle as the environment.  

To train the question generator part of the agent, we adopt REINFORCE~\cite{Williams92simplestatistical}, one of the on-policy policy gradient methods, because it scales well to large action spaces and can be applied to partially observable environments such as our case. Since the GuessWhat?! game is an episodic game, the objective is to find a question generation policy $\pi_\theta(q_t|o_t)$ that maximizes the expected cumulative reward:
\vspace*{-0.5\baselineskip}
\begin{equation}
\label{eq:pg}
J(\theta)=\mathbb{E}_{\pi_\theta}[\sum_{t=1}^T r(x_t,q_t)] 
=\mathbb{E}_{\pi_\theta}[R(\tau)].
\end{equation}
\vspace{-0.05cm}
The policy $\pi_\theta(q_t|o_t)$ can be improved by performing gradient ascent~\cite{Williams92simplestatistical}. The details are presented in Algorithm~\ref{algo:1} (see the REINFORCE part).

\subsection{Implementation}
\textbf{Question Generator.}
We model the question generator $\pi_\theta(q_t|o_t)$ by a 5-layer fully connected neural network conditioned on the concatenation of VGG16 FC8 feaures of the image $\mathcal{I}$ and the last hidden state of a LSTM encoding the QA history $h_{1:t-1}$, as illustrated in Fig.~\ref{fig:nn} (a).

\noindent
\textbf{Oracle Network.}
We adopt the same oracle network as proposed in~\cite{end_to_end_gw}, where the concatenation of the spatial and categorical information of the target object together with the last state of the LSTM which encodes the question is fed into a 2-layer fully connected neural network, as illustrated in Fig.~\ref{fig:nn} (b). The test error of our oracle implementation is 21.33\%. Note that with such an implementation the answer is conditionally independent of the previous QA history and the image, i.e. $p(a_t|c,q_t,h_{1:t-1},\mathcal{I})=p(a_t|c,q_t)$. 

\begin{figure}[t]
\begin{minipage}[b]{.49\linewidth}
  \centering
  \centerline{\epsfig{figure=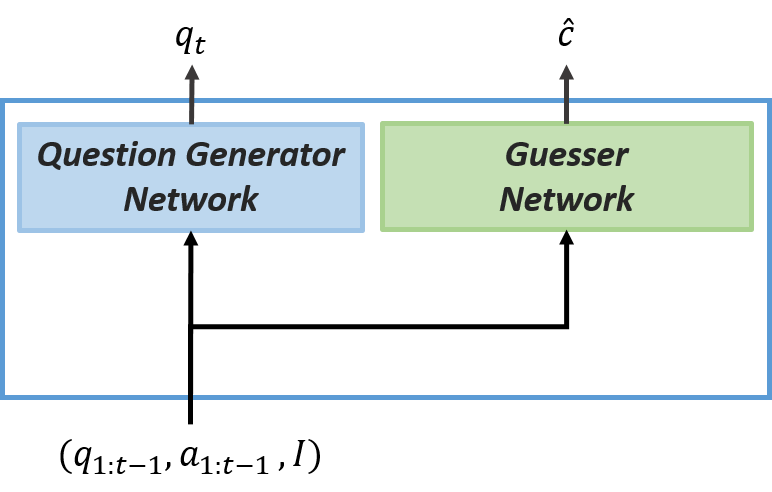,width=4.2cm}}
  \vspace{0.cm}
  \centerline{(a) Previous RL methods.}\medskip
\end{minipage}
\hfill
\begin{minipage}[b]{0.49\linewidth}
  \centering
  \centerline{\epsfig{figure=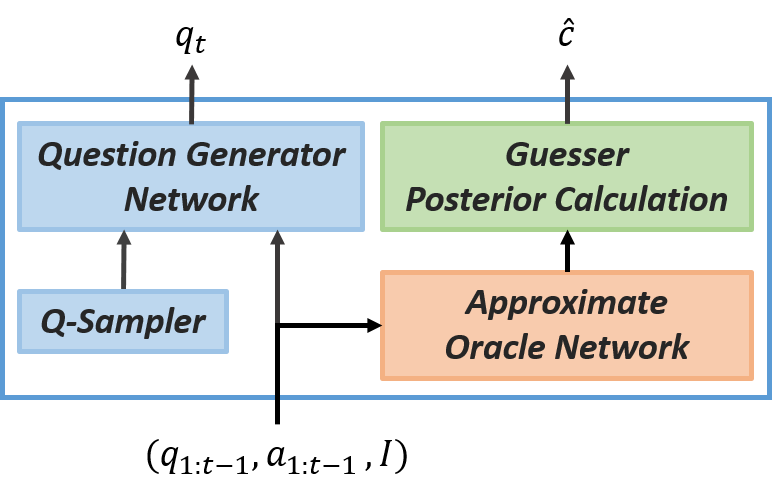,width=4.2cm}}
  \vspace{0.cm}
  \centerline{(b) Ours.}\medskip
\end{minipage}
  \vspace{-0.5cm}
\caption{Comparison of the questioner designs.}
\label{fig:nnn}
\vspace{-0.3cm}
\end{figure}


%


\begin{table*}[ht]

\caption{The testing prediction accuracy (in percentage) on GuessWhat?! dataset. Numbers are highlighted in boldface when our method outperforms IGE.} 

\resizebox{\textwidth}{!}{%

\begin{tabular}{|c|ccc|ccc|ccc|c|c|c|}
\hline
\multirow{2}{*}{} & \multicolumn{3}{c|}{trueA} & \multicolumn{3}{c|}{depA} & \multicolumn{3}{c|}{indA} & \multirow{2}{*}{Strub et al.~\cite{end_to_end_gw}} & \multirow{2}{*}{Zhao et al.~\cite{Zhao2018ImprovingGV}} & \multirow{2}{*}{Human~\cite{end_to_end_gw}} \\ \cline{2-10}
 & \multicolumn{1}{l}{AQM~\cite{NIPS2018_7524}} & IGE & Ours & AQM~\cite{NIPS2018_7524} & IGE & Ours & AQM~\cite{NIPS2018_7524} & IGE & Ours &  &  &  \\ \hline
1q & - & 32.78 & 31.92 & - & 31.72 & 31.39 & - & 32.29 & 31.15 & \multirow{10}{*}{58.4} & \multirow{10}{*}{74.31} & \multirow{10}{*}{84.4} \\
2q & - & 56.26 & 48.25 & - & 54.71 & 47.58 & - & 52.83 & 45.65 &  &  &  \\
3q & 63.76 & 70.71 & 60.18 & 63.63 & 70.23 & 55.73 & - & 63.91 & 54.26 &  &  &  \\
4q & - & 77.32 & 68.27 & - & 76.60 & 67.77 & - & 67.94 & 64.27 &  &  &  \\ \cline{1-10}
5q & - & 80.00 & 69.32 & 72.89 & 78.58 & 68.65 & 66.73 & 69.77 & 66.95 &  &  &  \\
6q & - & 81.03 & 75.86 & - & 79.22 & 75.55 & - & 70.40 & \textbf{71.52} &  &  &  \\
7q & - & 81.56 & 78.07 & - & 79.29 & 77.26 & - & 70.94 & \textbf{73.60} &  &  &  \\
8q & - & 81.79 & 80.96 & - & 79.44 & 79.13 & - & 71.15 & \textbf{74.85} &  &  &  \\
9q & - & 82.09 & 81.65 & - & 79.71 & \textbf{80.49} & - & 71.26 & \textbf{75.83} &  &  &  \\
10q & 81.96 & 82.24 & \textbf{82.45} & 78.72 & 79.71 & \textbf{80.68} & - & 71.43 & \textbf{76.31} &  &  &  \\ \hline
\end{tabular}%
}

\label{tab:NP_acc}
\vspace{-0.4cm}
\end{table*}


\section{Experiment}\label{sec:experiment}
This session compares the proposed method with AQM, IGE, and few other state-of-the-art baselines on the GuessWhat?! dataset, in terms of prediction accuracy. We follow the settings in~\cite{NIPS2018_7524} to test the robustness of different methods to the oracle approximation error. We conclude with a subjective evaluation. 

\subsection{Settings}
\noindent
\textbf{Dataset.}
GuessWhat?! dataset consists of 155K dialogues including 822K QA pairs on 67K unique images and 134K unique objects. The dataset is split randomly by assigning 70\%, 15\% and 15\% of the images and their corresponding dialogues into the training, validation and test set, respectively.

\noindent
\textbf{Q-Sampler.}
We use the countQ strategy described in~\cite{NIPS2018_7524} to sample 200 questions from the training data into $Q$. These questions are selected to be independent of each other in the sense that the probability of the answers to any two distinct questions being identical is no more than 0.95. We additionally add an extra requirement that the selected questions should appear at least 3 times in the training data. 

\noindent
\textbf{Approximate Oracle.}
To investigate how the non-ideal oracle model adopted by the questioner would affect prediction accuracy, we follow~\cite{NIPS2018_7524} to conduct experiments for 3 different approximate oracles: (1) indA, which is trained from the training data, (2) depA, which is trained from the images and questions in the training data together with answers given by the true oracle, and (3) trueA, which is a copy of the true oracle. Recall that the true oracle is a model of the answerer. The depA simulates the case where the questioner builds a model of the environment, i.e. the answerer, by learning online through interacting with the answerer.

\noindent
\textbf{Reward.}
During RL training, the questioner gets no immediate rewards until it’s guesser successfully locates the target object at the end of T QA rounds, where a reward 1 is given.

\vspace{-0.13cm}

\subsection{Training Details}
\textbf{Progressive Training.} 
For training, we first train the questioner with IL for several iterations and then refine it with RL. By observing that IGE's performance begins to saturate in the \nth{5} QA round, we have the questioner imitate the analytic experts in the first 5 QA rounds before entering the RL phase.
In RL training, separate questioners are obtained progressively for QA rounds ranging from $T=5$ to $T=10$. 
That is, the questioner to be trained for $T$ QA rounds is initialized by the parameters that have already been trained for $T-1$ QA rounds.

\noindent
\textbf{Mixture of Experts.} 
Motivated by the intuition that human players tend to ask dividing, or binary, questions such as ``Is it in the left side of the image?'' in the early rounds of the game followed by more specific questions such as ``It is the \nth{3} blue vase counting from the left?'' after they acquire some confidence about the target object, we arrange for the questioner to mimic IGE in the first 4 rounds and TPE in the \nth{5} round. 
Compared to using IGE alone in all 5 rounds, this design with a mixture of experts turns out to give a 4.31\% boost to the validation prediction accuracy after IL.
The performance deteriorates too when the questioner is allowed to mimic TPE's actions in early rounds. Since TPE exploits the information of the target object, it tends to ask identifying questions that directly relate to it, resulting in an enumerating strategy. For example, if the target object is an apple, TPE may select ``is it an apple?'' as its very beginning question, which becomes linear search and is generally not a good strategy.


\begin{table*}[ht]

\caption{Qualitative comparison of the dialogues generated by IGE and our method in depA setting.} 

\resizebox{\textwidth}{!}{%
\begin{tabular}{llrlrlr}
\hline
\multicolumn{1}{c}{Image} & \multicolumn{2}{c}{Ground Truth} & \multicolumn{2}{c}{IGE} & \multicolumn{2}{c}{Ours} \\ \hline
\multirow{11}{*}{
\raisebox{-.5\height}{\includegraphics[width=4cm]{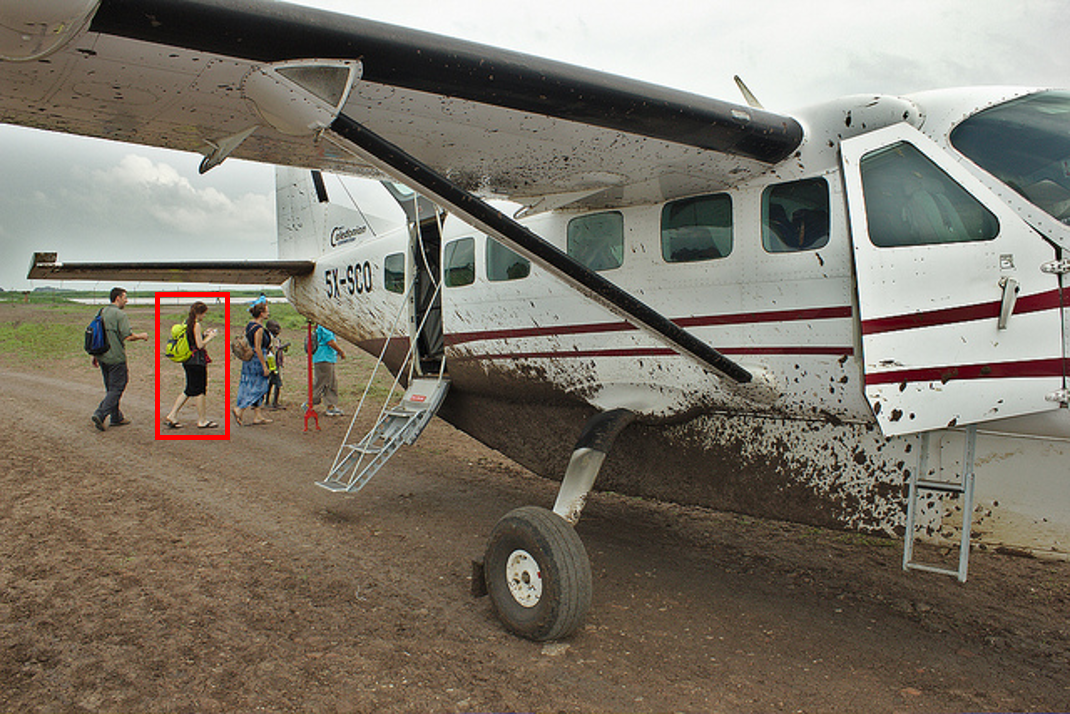}}
} & A person? & Yes. & A person? & Yes. & A person? & Yes. \\
 & The left most? & No. & The furthest left? & No. & Is in right side? & No. \\
 & The 2nd from left to right? & Yes. & Boy on left? & Yes. & Is it on the left side of picture? & Yes. \\
 &  &  & The furthest left? & No. & One of the two closest to us? & No. \\
 &  &  & Boy on left? & Yes. & On the bottom row? & Yes. \\
 &  &  & Boy on left? & Yes. & On top shelf? & No. \\
 &  &  & Boy on left? & Yes. & I in middle? & No. \\
 &  &  & The furthest left? & No. & Boy on left? & Yes. \\
 &  &  & Boy on left? & Yes. & Is it near the left edge of the photo? & No. \\
 &  &  & Boy on left? & Yes. & Is it near the left edge of the photo? & No. \\
 & Status: & \color{ForestGreen} Success. & Status: & \textcolor{red}{Failure.} & Status: & \color{ForestGreen} Success. \\ \hline
\end{tabular}%
}

\label{tab:qua}
\vspace{-0.3cm}
\end{table*}


\vspace{-0.12cm}

\subsection{Experimental Results}
Table \ref{tab:NP_acc} presents the quantitative comparison results with other state-of-the-art works in terms of testing prediction accuracy.
In our proposed method, all the questioners with different approximate oracle settings are pretrained for 5 QA rounds by IL and then refined by RL progressively for QA rounds ranging from $T=5$ to $T=10$. For completeness, we also train questioners by RL after IL without progressive training for QA rounds ranging from $T=1$ to $T=4$.

From Table \ref{tab:NP_acc}, several observations can be made. First, it is worth noting that at test time, IGE has access to the ground-truth list of candidate objects, thereby having better knowledge of the posterior $p(c|\mathcal{I})$. This explains why it outperforms AQM, which relies on a non-ideal object detector to initialize the posterior. Due to their algorithmic similarity, we consider IGE to be a stronger AQM-like baseline. 

Second, when provided with the same approximate oracle, our questioner successfully surpasses its expert IGE in games with more QA rounds.
More specifically, our method intersects and outperforms IGE in games with 10, 9, 6 QA rounds when the approximate oracle is trueA, depA and indA, with their respective success rates being from 82.45\% to 82.24\%, 80.49\% to 79.71\%, and 71.52\% to 70.40\%. This confirms that the RL training phase indeed discovers a better policy than the expert's.

Third, it is seen that our method is more robust to the approximation error of the oracle. Recall from Section \ref{sec:ILExperts} that IGE needs an approximate oracle model to evaluate the information gain and that the same model will also be used by both IGE and our method to make a guess in the final QA round. We observe that the prediction accuracy of IGE drops in the 10-round game by as much as 10.81\%, from 82.24\% with trueA to 71.43\% with indA. In contrast, our method largely mitigates this performance decline to 6.14\%, suggesting that through the refinement of RL, our model can better accommodate the discrepancy between the true oracle and the approximate oracle in the questioner.

Lastly, we see that by planning only one step, IGE can already achieve closely human-level performance. That is why our method benefits only moderately from RL, the advantage of which is most obvious in solving problems requiring long-sequence, dependent decision making.

A taste of the dialogues generated by IGE and our method is given in Table 2. We see that IGE sticks to certain questions while our model asks different questions to successfully locate the target object. However, we also observe that our model makes heavy use of questions about relative positions and directions among most of the games, which may provide clues to why it needs more QA rounds to outperform IGE.

%

\section{Conclusion} \label{conclusion}
 
We train a questioner for the GuessWhat?! task based on imitation and reinforcement learning. We develop two analytic experts, IGE and TPE, for imitation learning on top of the probabilistic framework developed for AQM. Because both experts are greedy and have high reliance on an accurate oracle model of the answerer, we further refine our model using the REINFORCE algorithm. Our method outperforms the conventional RL-based models trained with limited expert demonstrations by a large margin, while surpassing IGE expert in games with long QA rounds in terms of prediction accuracy and robustness. 
To develop a questioner that is able to outperform analytic experts in games with short QA rounds and to test the performance of our method on more challenging datasets are among the scope of future works.

\vspace{-0.055cm}
%

\bibliographystyle{IEEEbib}
\bibliography{gw_ref}

\end{document}